\documentclass[a4paper,fleqn]{cas-dc}

\usepackage[authoryear,longnamesfirst]{natbib}
\usepackage{caption}
\def\tsc#1{\csdef{#1}{\textsc{\lowercase{#1}}\xspace}}
\tsc{WGM}
\tsc{QE}

\begin{document}
\let\WriteBookmarks\relax
\def\floatpagepagefraction{1}
\def\textpagefraction{.001}

\shorttitle{Smartphone-based Circular Plot Sampling for Forest Inventory}    

\shortauthors{Su Sun et al.}  

\title [mode = title]{Smartphone-based Circular Plot Sampling for Forest Inventory}  



\author[1]{Su Sun}

\ead{sun931@purdue.edu}

\credit{Conceptualization, Methodology, Software, Validation, Formal analysis, Investigation, Resources, Data curation, Writing -- original draft, Visualization}

\author[1]{Jui-Cheng Chiu}

\ead{chiu119@purdue.edu}

\credit{Data curation, Writing -- original draft, Writing -- review \& editing}

\author[1]{Nabin Khanal}

\ead{khanaln@purdue.edu}

\credit{Data curation, Writing -- original draft, Writing -- review \& editing}

\author[2]{Songlin Fei}[orcid=0000-0003-2772-0166]

\ead{sfei@purdue.edu}

\credit{Conceptualization, Methodology, Writing -- review \& editing, Supervision, Project administration, Funding acquisition}

\author[1]{Yingjie Victor Chen}

\cormark[1]

\ead{victorchen@purdue.edu}

\credit{Conceptualization, Methodology, Writing -- review \& editing, Supervision, Project administration, Funding acquisition}

\affiliation[1]{organization={School of Applied and Creative Computing, Purdue University},
            addressline={}, 
            city={West Lafayette},
            postcode={}, 
            state={IN},
            country={USA}}

\affiliation[2]{organization={Department of Forestry and Natural Resources, Purdue University},
            addressline={}, 
            city={West Lafayette},
            postcode={}, 
            state={IN},
            country={USA}}

\cortext[1]{Corresponding author}

\begin{abstract}
Circular sample plots are a cornerstone of forest inventory, yet accurate measurement of tree diameter at breast height (DBH) and spatial location within such plots remains challenging. Conventional approaches rely either on costly terrestrial LiDAR systems or labor-intensive manual methods involving calipers and compass bearings, limiting their scalability and accessibility in large scale environments. We present a lightweight, smartphone-based pipeline that enables complete plot sampling based tree measurement from a single walkthrough video, requiring no specialized hardware beyond a consumer smartphone mounted on a portable stand. The proposed method integrates pretrained monocular depth estimation and tree instance segmentation with a simultaneous localization and mapping (SLAM) framework to jointly refine camera trajectories and depth across the video sequence. Tree positions and DBH estimates are recovered by fusing SLAM-derived camera poses with segmented depth maps, with absolute real-world scale anchored via a calibrated reference length. 

The system was evaluated in both managed forest plots and natural forest plot, achieving a mean absolute error of 1.51 cm (MARE 3.98\%) and 2.30 cm (MARE 5.69\%) respectively, with consistent performance across varying starting directions and positions.
Cross-video consistency analysis further demonstrated stable and reproducible tree localization across measurements initiated from different starting positions. The proposed approach achieves accuracy comparable to established field methods while substantially reducing equipment cost and operational complexity, making it accessible to both professional researchers and non-expert forest managers in diverse operational settings.

\end{abstract}




\maketitle

\section{Introduction}\label{sec:introduction}
Forest inventories provide the foundation for timber management, carbon stock assessment, biodiversity monitoring, and climate change mitigation policy. One common inventory method is circular plot sampling: a fixed-radius area within which field crews measure individual tree attributes such as diameter at breast height (DBH), height, and species composition. DBH is arguably the most fundamental of these attributes, serving as the primary input for estimating basal area, stem volume, above-ground biomass, and carbon stock at the stand level. In operational settings, these measurements are commonly obtained using manual instruments such as diameter tapes, calipers, compasses, and clinometers. Although such tools are inexpensive and well established, their applications remain labor-intensive and scale poorly across large forest areas. These limitations have motivated sustained interest in sensing technologies capable of improving efficiency while maintaining plot-level measurement accuracy.

Among existing technologies, terrestrial laser scanning (TLS) has emerged as a new digital tool for detailed plot-level forest measurement. TLS systems produce dense three-dimensional point clouds that enable accurate estimation of stem position, DBH, tree height, and taper with high geometric precision~\citep{calders2020tls,liang2019benchmark, stovall2023stemtaper}. However, TLS remains constrained by high equipment cost, multi-scan acquisition requirements, occlusion sensitivity, and substantial post-processing overhead~\citep{newnham2015tls}. Hand-held and backpack mobile laser scanning systems improve portability and simplify field operation~\citep{holopainen2022handheld_pls,tao2021tls_largearea}.


Recent work further demonstrates the effectiveness of mobile laser scanning for forest inventory and stem structure estimation~\citep{shao2026mls,hanafy2025lidar}. Nevertheless, these systems still rely on active ranging hardware that is often inaccessible to small forest operations and resource-limited regions.


To improve accessibility, recent work has explored consumer-grade mobile devices as forest inventory tools. Smartphone-based image acquisition and photogrammetric reconstruction have been shown to produce DBH estimates comparable to manual field measurements under favorable conditions~\citep{marzulli2020smartphone,ahamed2023_phone_photogrammetry,li2020smartphone,xiang2025smartphone}. These approaches build upon broader developments in structure-from-motion (SfM) and multi-view stereo (MVS) reconstruction from consumer imagery~\citep{snavely2006photo,schonberger2016sfm,furukawa2015mvs}. More recently, LiDAR-equipped mobile devices such as modern iPhones and iPads have further expanded the possibilities of lightweight forest sensing. Applications such as iForester \citep{Yin}  and ForestScanner \citep{tatsumi2023forestscanner} demonstrate that embedded LiDAR sensors can rapidly measure and map trees in the field, and several studies report encouraging agreement with conventional DBH measurements~\citep{howie2024_ipad_lidar_dbh,gulci2023smartphone}. Despite these advances, LiDAR-enabled approaches remain tied to specific hardware generations and are therefore not universally available. Moreover, many existing mobile solutions still operate in a tree-by-tree interaction paradigm, requiring explicit operator attention to each individual stem, which limits their efficiency for full circular plot inventory.

A more scalable approach is to reconstruct the geometry of the entire plot from a single continuous video capture and extract all tree measurements from the resulting three-dimensional reconstruction. This desired approach naturally aligns with simultaneous localization and mapping (SLAM), which jointly estimates camera trajectory and scene structure from image sequences. Classical feature-based SLAM systems such as ORB-SLAM2~\citep{mur2017orb} and direct methods such as Direct Sparse Odometry (DSO)~\citep{engel2018direct} have demonstrated strong performance in structured environments. Foundational developments in visual SLAM and visual odometry~\citep{cadena2016slam,scaramuzza2011visual,klein2007ptam} underpin these approaches. 


However, forest scenes present several challenges, including repetitive bark textures, foliage clutter, heavy self-occlusion, illumination variability, and limited geometric regularity, which degrade feature matching and reconstruction stability~\citep{garforth2019_forest_slam}. Recent learning-based and dense SLAM systems such as DROID-SLAM~\citep{teed2021droid}, Co-SLAM~\citep{wang2023coslam}, and HI-SLAM2~\citep{zhang2025hislam2} improve tracking robustness and dense reconstruction quality. Recent forest-oriented mapping frameworks further enhance trajectory estimation and fine-scale forest reconstruction using mobile sensing platforms~\citep{zhao2026is2team}. Nevertheless, robust generalization to unconstrained outdoor forest environments remains challenging.

In addition to tracking instability, monocular SLAM suffers from inherent scale ambiguity: without external references, the reconstructed scene is defined only up to an unknown scale factor. This limitation prevents direct extraction of metric quantities such as DBH. Prior work attempting monocular reconstruction for forestry applications~\citep{piermattei2019improving} has therefore relied on simplified models or manual scale initialization, limiting practical deployment.

Meanwhile, rapid advances in deep learning–based monocular depth estimation have produced models capable of generating dense scene geometry from a single RGB image. Early approaches such as Eigen et al.~\citep{eigen2014depth} demonstrated the feasibility of learning-based depth prediction, while later models such as DPT and MiDaS significantly improved cross-dataset generalization~\citep{ranftl2021_dpt}. Recent work such as Depth Anything further advances generalization through large-scale training~\citep{li2024_depth_anything}, while Video Depth Anything introduces temporal consistency for video-based applications~\citep{zhao2024_video_depth_anything}. These developments provide an opportunity to directly estimate metric depth from monocular inputs, but with limited applications in forest environments \citep{Lee}.

Despite extensive research on TLS-based forest measurement, mobile sensing applications, and visual SLAM, a practical solution for full circular sample plot inventory using only widely available consumer hardware remains lacking. Existing LiDAR-based mobile systems depend on specialized hardware and often require tree-by-tree scanning, while classical SLAM methods struggle with the visual complexity of forest environments and cannot recover metric measurements without external scale references. Smartphone photogrammetry approaches typically focus on individual trees or require multi-view reconstruction workflows rather than continuous video-based capture. Consequently, there remains a need for a lightweight system capable of reconstructing forest plots and estimating DBH for stems using only monocular video captured with standard smartphones.

In this paper, we present a lightweight, smartphone-based system for measuring all trees within a circular sample plot from a single handheld video, requiring no equipment beyond a consumer smartphone mounted on a low-cost stand. Our approach is built around a circular data collection protocol in which the user walks around a fixed plot center with the camera pointing outward, completing a full 360\textdegree{} rotation in around one minute. This trajectory is deliberately designed to provide natural loop-closure opportunities, a known physical radius for metric scale recovery, and consistent viewpoint coverage of stems in the plot. Our reconstruction pipeline integrates a video depth model as a soft structural prior within a dense visual SLAM framework, stabilizing tracking in low-texture forest environments without requiring model fine-tuning on forest-specific data. Tree stems are identified and isolated via semantic segmentation, and DBH is estimated through a projection-based cross-sectional analysis that is robust to depth noise and partial occlusion. Metric scale is recovered from the known stand radius, requiring no external sensors or calibration targets. We validate the system in both planted and natural environment in Martell Forest at Purdue University, West Lafayette. Cross-video consistency analysis is also conducted. 

Our main contributions are as follows:



\begin{itemize}
    \item We introduce a circular acquisition protocol using a portable tripod-mounted stand that mechanically stabilizes the camera trajectory to reduce motion jitter, improving reconstruction quality while enabling natural loop closure, consistent stem coverage, and metric scale recovery from a single known reference length without external calibration targets.
    \item We propose a complete, smartphone-based pipeline for plot-level forest inventory that measures tree DBH and position for stems within a circular sample plot from a single handheld monocular video, requiring no specialized hardware or expert operation.
    \item We demonstrate that a large pre-trained video depth model can be effectively integrated as a soft regularization prior within a dense monocular SLAM framework, improving tracking robustness and reconstruction quality in low-texture wild forest environments.
    \item We propose a segmentation-guided, projection-based DBH estimator using reconstructed point clouds, achieving a MARE of 3.98\% across four managed forest plots and 5.69\% on a natural forest plot, comparable to existing smartphone-based and lightweight LiDAR-based approaches.

\end{itemize}

\section{Materials and Methods}\label{sec:methods}
In this section, we describe the full technical pipeline of our system, using video-based 3D reconstruction for forestry inventory.

\subsection{System Overview}

\begin{figure*}[ht!]
\centering
\includegraphics[width=.9\textwidth]{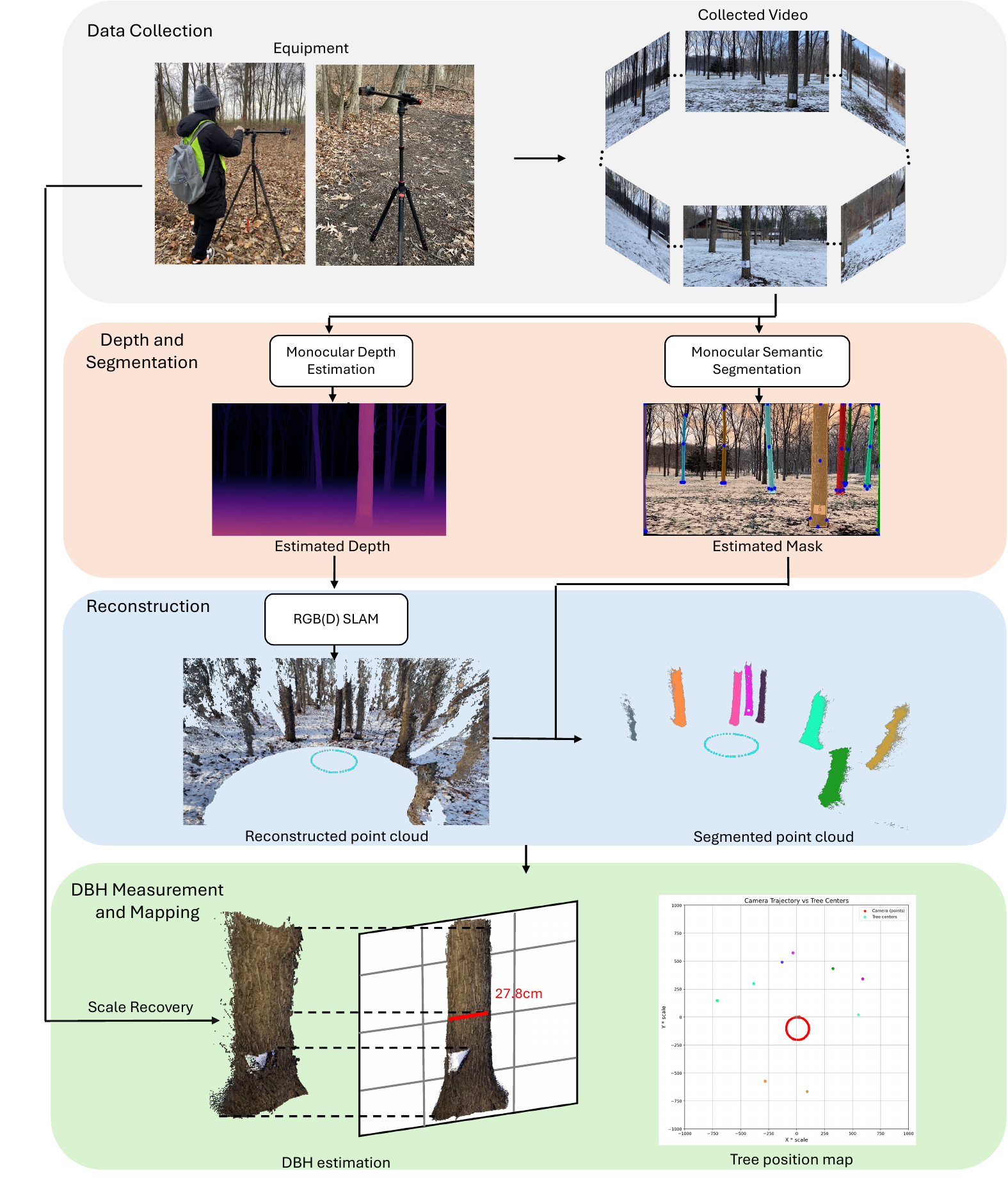}
\caption{
Overview of the proposed smartphone-based forest inventory pipeline.
A monocular video is captured by rotating a smartphone-mounted stand
within a circular sample plot (\emph{top}). Dense depth estimation and
semantic segmentation are performed for each frame (\emph{upper middle}).
A SLAM module refines the prior depth and reconstructs a globally consistent 3D point cloud
and estimates camera poses (\emph{lower middle}). Tree stems are then
clustered based on segmentation masks to estimate DBH, producing a tree
position map. The real-world scale is recovered by the radius of the phone stand (\emph{bottom}).
}
\label{fig:pipeline_overview}
\end{figure*}

Our objective is to estimate tree positions and diameters from a single handheld monocular video. Given an input sequence $\{I_i \in \mathbb{R}^{H \times W \times 3}\}_{i=0}^{N}$, our system maintains for each frame a camera pose $G_t \in SE(3)$ and an inverse depth map $d_i \in \mathbb{R}^{H \times W}$, which together yield a metrically consistent 3D reconstruction. Within the reconstruction, each tree stem indexed by $k$ is represented by a 3D position $\mathbf{p}_k \in \mathbb{R}^3$ in the global coordinate frame, and its diameter at breast height is denoted $\mathrm{dbh}_k$.

Our system takes as input a short monocular video captured by rotating the camera 360\textdegree{} around the target area. This acquisition strategy provides natural loop-closure opportunities and a roughly constant scanning radius, both of which are exploited in later stages. The pipeline combines three core components: (1) a pre-trained monocular depth model that provides dense depth priors to stabilize reconstruction in low-texture forest environments; (2) a deep visual SLAM system that estimates camera trajectories and recovers consistent scene geometry; and (3) a segmentation-guided measurement module that identifies individual tree stems in the reconstructed map and estimates their positions and diameters. The following subsections describe each component in detail.

\subsection{Video Collection}
To acquire input data, the user mounts a smartphone on a simple rotation stand placed at the center of the sample plot. The stand consists of a tripod base and a horizontal arm extending approximately one meter from the rotation axis, with the smartphone attached at the far end. Compared to handheld capture, the rigid stand serves two purposes: it mechanically constrains the camera to a smooth circular trajectory, reducing the irregular motion jitter that would otherwise degrade SLAM tracking, and it fixes the scanning radius to a known physical length that is later used for metric scale recovery. The user pushes the opposite end of the arm to move the camera along a circular trajectory of approximately one-meter radius while continuously facing the surrounding trees. The stand is positioned near the center of a circular sample plot. The camera is mounted at breast height (approximately 1.3\,m above ground) to ensure that the standard DBH measurement zone remains within the image region throughout the scan.


A single complete rotation takes 40--50 seconds and does not require precise speed control; any modern smartphone recording 1080p video at 30\,FPS is sufficient. To verify completeness, the operator selects a reference tree at the start of recording and continues rotating until the camera returns to the same tree from the opposite side, confirming a full 360\textdegree{} scan. The circular trajectory benefits the downstream pipeline in three ways: (i) the rigid stand stabilizes the camera trajectory, mitigating motion jitter that degrades feature tracking and depth estimation; (ii) the natural return to the starting viewpoint provides reliable loop-closure constraints that suppress accumulated drift; and (iii) the known physical radius of the stand anchors metric scale recovery without external calibration targets. No specialized sensors (GPS, IMU, or LiDAR) are required, and the entire acquisition process completes in under two minutes per plot.

\subsection{Dense Monocular Depth Prior}
Monocular SLAM suffers from inherent scale ambiguity and unstable tracking, particularly in environments with weak or repetitive textures. Forest scenes present a challenging case: tree trunks, bark, and understory vegetation often lack distinctive feature points required by conventional feature-based SLAM systems. As a result, homogeneous or repeated textures produce sparse and unreliable correspondences, which may lead to tracking failures and inaccurate pose estimation. Dense monocular depth estimation provides complementary geometric cues that can alleviate these limitations by supplying per-pixel structural information, enabling more stable reconstruction in feature-poor regions.

Early large-scale monocular depth models such as MiDaS and DPT demonstrated strong cross-dataset generalization \cite{ranftl2021_dpt}. However, these models operate on individual frames and do not enforce temporal consistency. When applied to video sequences, independent frame-wise predictions often introduce temporal instability, resulting in depth flickering and inconsistent geometry. Such artifacts are particularly problematic for SLAM systems, as they can introduce spurious depth discontinuities that degrade tracking and dense reconstruction.

Recent research has therefore focused on incorporating temporal information into depth estimation. Metric3D \cite{metric3d2023} leverages camera intrinsics to produce metric-consistent depth predictions. Depth Anything \cite{li2024_depth_anything} achieves strong generalization through large-scale training across diverse datasets. Most relevant to this work, Video Depth Anything (VDA) \cite{zhao2024_video_depth_anything} improves temporal consistency by aligning motion-aware features and performing temporal smoothing across adjacent frames. This design significantly reduces frame-to-frame depth fluctuations and produces temporally coherent geometry, making VDA well suited as a structural prior for monocular SLAM in low-texture environments.

Despite these advantages, VDA and similar models are primarily trained on general-purpose datasets and are not optimized for forest scenes. Consequently, their predictions may exhibit scale ambiguity or frame-dependent offsets when applied to wild forest environments. Directly enforcing these raw depth values in SLAM optimization could therefore introduce systematic errors in both camera trajectories and reconstructed geometry. To address this issue, we incorporate the depth prediction as a soft prior rather than a hard constraint. This formulation allows the optimization to exploit the structural information provided by the depth network while remaining robust to scale and offset inaccuracies. For each input frame $I_i$, the VDA model produces a dense depth prediction $d_i^{*} \in \mathbb{R}^{H \times W}$. The integration of this prior into our SLAM optimization is detailed in the following subsection.

\begin{figure*}[!h]
\includegraphics[width=.9\textwidth]{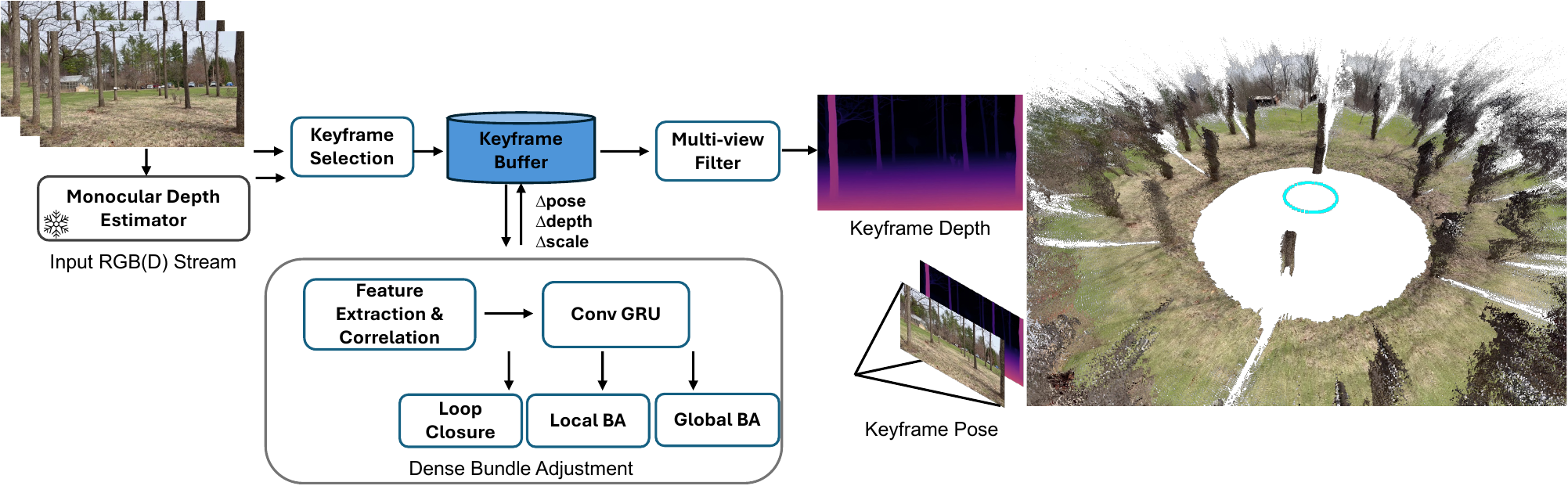}
\caption{End-to-end dense SLAM pipeline. A frozen monocular depth estimator supplies per-frame depth priors to a keyframe buffer. Dense correspondences are computed via feature extraction, correlation, and a convolutional GRU, which jointly updates poses, depth, and prior scale/offset through local and global bundle adjustment. A loop closure module injects long-range edges to suppress drift. \textit{Right}: recovered camera trajectory with the reconstructed point cloud.
\label{fig:slam_pipeline}}
\end{figure*}
\unskip

\subsection{End-to-End Dense SLAM with Loop Closure}

Our SLAM module builds upon the DROID-SLAM framework \cite{teed2021droid}, which performs dense visual SLAM through recurrent iterative updates of camera pose and per-pixel depth via a differentiable Dense Bundle Adjustment (DBA) layer. We extend this framework with a monocular depth prior integration and a visual loop closure mechanism.

\subsubsection{Frame Graph and Dense Correspondence}
The system maintains a frame graph $\mathcal{G} = (\mathcal{V}, \mathcal{E})$ over a set of keyframes, where each node $i \in \mathcal{V}$ is associated with a camera pose $G_i \in SE(3)$ and a dense inverse depth map $d_i \in \mathbb{R}^{H \times W}$. Edges $(i,j) \in \mathcal{E}$ encode co-visibility relationships between frame pairs. For each input image, a feature extraction network consisting of residual blocks and downsampling layers produces dense feature maps at $1/8$ the input resolution. For every edge $(i,j)$ in the graph, a 4D correlation volume is computed by taking the dot product between all pairs of feature vectors from the two frames, and a multi-scale correlation pyramid is constructed via average pooling. These correlation features, together with the induced optical flow and context features, are injected into a convolutional GRU that iteratively predicts (i) a dense correspondence revision field $r_{ij} \in \mathbb{R}^{H \times W \times 2}$ and (ii) a per-pixel confidence map $w_{ij} \in \mathbb{R}^{H \times W \times 2}_+$.

The corrected correspondences $\hat{p}_{ij} = r_{ij} + p_{ij}$ are compared against the geometric reprojection of pixels from frame $i$ into frame $j$. For an edge $(i,j) \in \mathcal{E}$, a pixel $p_i$ in frame $i$ is lifted to 3D via inverse projection and reprojected into frame $j$:
\begin{equation}
    p_{ij} = \Pi_c\!\left(G_{ij}\,\Pi_c^{-1}(p_i, d_i)\right), \qquad G_{ij} = G_j G_i^{-1},
\end{equation}
where $\Pi_c$ denotes the camera projection model and $\Pi_c^{-1}$ its inverse. The discrepancy between the learned and geometric correspondences forms the dense reprojection objective that drives joint optimization of all poses and depth maps.

\subsubsection{Monocular Depth Prior Integration}
Directly enforcing the raw depth prior $d_i^{*}$ is problematic because monocular depth predictions are typically biased in scale and may carry additive offsets that vary across scenes and lighting conditions. We associate each keyframe $i$ with a learnable scale $s_i$ and offset $o_i$ that align $d_i^{*}$ with the optimized inverse depth $d_i$. The resulting depth prior residual is:
\begin{equation}
    r_i^{\text{depth}} = \big\|\, s_i \cdot d_i^{*} + o_i - d_i \,\big\|_2^2.
\end{equation}
This formulation isolates per-frame calibration errors in the monocular predictor from the underlying multi-view geometry. To resolve the inherent gauge ambiguity among scale, offset, and pose variables, we employ an alternating optimization scheme that first updates camera poses with alignment parameters held constant, then refines depth, scale, and offset estimates with the pose graph fixed.

\subsubsection{Joint Optimization}
Our joint optimization minimizes both the dense reprojection error and the depth prior residuals:
\begin{equation}
    \min_{G,d,s,o} \;
    \sum_{(i,j)\in\mathcal{E}}
    \big\|\hat{p}_{ij} - \Pi_c\!\left(G_{ij}\,\Pi_c^{-1}(p_i, d_i)\right)\big\|_{\Sigma_{ij}}^{2}
    \;+\;
    \lambda_d \sum_{i\in\mathcal{V}} r_i^{\text{depth}},
\end{equation}
where $\Sigma_{ij} = \mathrm{diag}(w_{ij})$ denotes the per-correspondence confidence weights predicted by the GRU and $\lambda_d$ controls the strength of the depth prior. The reprojection term enforces multi-view geometric consistency through learned dense correspondences, while the depth regularization term anchors the solution to the structure predicted by the monocular model. Together, these objectives enable robust reconstruction in low-texture, repetitive forest environments where purely geometric SLAM methods are prone to failure or noisy geometry.

Following DROID-SLAM \cite{teed2021droid}, the optimization is split into a local frontend that performs bundle adjustment over a sliding window of recent keyframes and a global backend that periodically re-optimizes the full trajectory to enforce long-range consistency.

\subsubsection{Loop Closure}
Because our circular scanning trajectory causes the camera to naturally revisit similar viewpoints, we exploit these opportunities for loop closure to suppress long-range drift. We adopt learned place recognition features \cite{Berton_2023_EigenPlaces} for loop candidate retrieval, which generalize more reliably than hand-crafted descriptors in forest scenes. For each incoming keyframe $i$, we extract a global visual descriptor $\mathbf{f}_i$ and query it against an index of all previous keyframes. A past frame $j$ is accepted as a loop candidate only if the descriptors are sufficiently similar and the two frames are temporally far apart, preventing false positives from consecutive frames with overlapping fields of view.
Each validated pair $(i, j)$ yields a relative pose estimate $\hat{G}_{ij}$ via the existing dense correspondence mechanism.

Accepted loop edges are inserted into the global backend graph, and loop consistency is enforced by an additional constraint over the loop edge set $\mathcal{L}$:
\begin{equation}
    \mathcal{E}_{\text{loop}}
    = \sum_{(i,j)\in\mathcal{L}}
    \left\|\log\!\left(\hat{G}_{ij}^{-1} G_j G_i^{-1}\right)\right\|^2,
\end{equation}
where $\log(\cdot)$ denotes the $\mathfrak{se}(3)$ logarithm. This term penalizes inconsistencies between the loop-edge relative pose and the current trajectory, anchoring the reconstruction at revisited viewpoints. Accepted loop edges are incorporated into the global backend, correcting accumulated drift across the full trajectory. The circular video collection method guarantees at least one strong loop closure per scan, substantially improving global consistency of both camera poses and the reconstructed geometry.

\subsection{Segmentation-Based Tree Measurement}

\subsubsection{Point Cloud Generation and Stem Detection}
To extract tree measurements from the reconstructed scene, we first obtain a semantic segmentation mask $\mathcal{M}_i \in \{0,1\}^{H \times W}$ for each frame using a pretrained segmentation model. Masked pixels are back-projected into 3D using the estimated inverse depth $d_i$ and camera pose $G_i$:
\begin{equation}
    \mathcal{P}_i =
    \Big\{ G_i \, \Pi_c^{-1}(p, d_i(p)) \;\Big|\; p \in \Omega,\; \mathcal{M}_i(p) = 1 \Big\}.
\end{equation}
Aggregating these sets across all frames yields a dense, though noisy, forest point cloud.

Individual stem candidates are identified by tracking frames in which a tree appears near the image center. For each such frame $i$, the back-projected 3D point of the central pixel is treated as a tentative stem center $\mathbf{c}_i$. After transforming all candidates into the global frame, we retain only those within a 10\,m radius of the trajectory centroid to remove distant stems with insufficient observational support. The remaining candidates are clustered in 3D using DBSCAN, yielding a set of candidate stem clusters $\{\mathcal{C}_k\}$.

Unreliable clusters are pruned using two criteria. First, we apply PCA to each cluster and measure the alignment of its dominant eigenvector $\mathbf{e}_1$ with the world vertical axis $\mathbf{v}_z$ via $s_k = |\mathbf{e}_1^\top \mathbf{v}_z|$; clusters below a threshold $\tau_s$ are discarded as non-vertical. Second, clusters with insufficient vertical extent $h_k = \max z(\mathcal{C}_k) - \min z(\mathcal{C}_k)$ are removed. Together, these filters retain coherent, vertically elongated structures corresponding to tree stems while eliminating short or noisy outlier clusters.

\subsubsection{DBH Estimation}
\begin{center}
\includegraphics[width=.8\columnwidth]{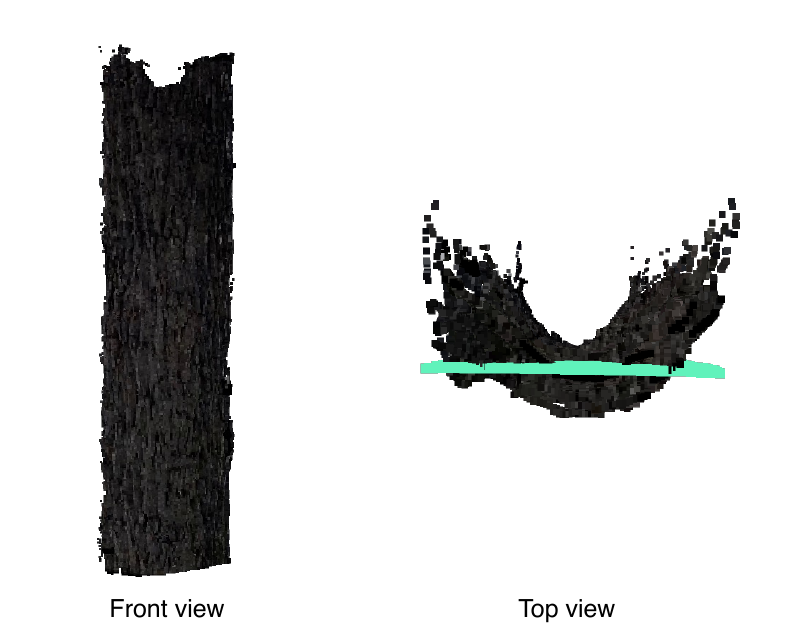}
\captionof{figure}{Reconstructed point cloud of an individual tree stem. \textit{Left}: front view showing dense coverage on the camera-facing surface. \textit{Right}: top-down view illustrating the partial observation due to the circular scanning trajectory; the cyan plane indicates the projection surface used for diameter estimation, oriented perpendicular to the camera-to-stem direction.\label{fig:projection}}
\end{center}

For each retained cluster $\mathcal{C}_k$, we estimate DBH using a cross-sectional slice analysis. We first extract a horizontal band of points at breast height (1.37\,m above ground) within a vertical window $\Delta h = 0.1$\,m.


Because the circular scanning trajectory only observes the front-facing surface of each tree, the reconstructed points are concentrated on the side nearest to the camera, with increasingly noisy and sparse coverage toward the lateral edges. To obtain a robust width estimate under this partial observation, we orthogonally project the slice points onto a plane perpendicular to the line connecting the trajectory center $\mathbf{o}$ to the stem center $\mathbf{c}_k$, with normal: 
\begin{equation}
    \mathbf{n}_k = \frac{\mathbf{c}_k - \mathbf{o}}{\|\mathbf{c}_k - \mathbf{o}\|}.
\end{equation}
This front-facing projection (Figure~\ref{fig:projection}) collapses noisy edge points onto the measurement axis where observation density is highest, reducing the influence of unreliable lateral geometry.
Projected points are expressed in a local orthonormal basis $(\mathbf{w}_k, \mathbf{h}_k)$, where $\mathbf{w}_k$ is the world vertical direction projected onto the plane and $\mathbf{h}_k = \mathbf{n}_k \times \mathbf{w}_k$.

\subsubsection{Metric Scale Recovery}
Because monocular SLAM reconstructs the scene up to an unknown scale factor, all 3D quantities are initially in a relative coordinate system. We recover metric scale by exploiting the known physical geometry of the data capture setup, as illustrated in Figure~\ref{fig:scale_recovery}. The camera is mounted on a stand at a known radius $r_{\text{real}}$ from the rotation axis. We estimate the corresponding reconstructed trajectory radius $r_{\text{est}}$ as the mean distance from each camera center $\mathbf{o}_i$ to the trajectory centroid $\bar{\mathbf{o}}$, and compute the global scale factor as:

\begin{equation}
    s = \frac{r_{\text{real}}}{r_{\text{est}}}.
\end{equation}
All reconstructed quantities including 3D point clouds, camera poses, and diameter estimates are then converted to metric units by applying $s$. This procedure provides reliable and repeatable metric scaling without requiring external sensors or calibration targets.

\begin{center}
\includegraphics[width=.9\columnwidth]{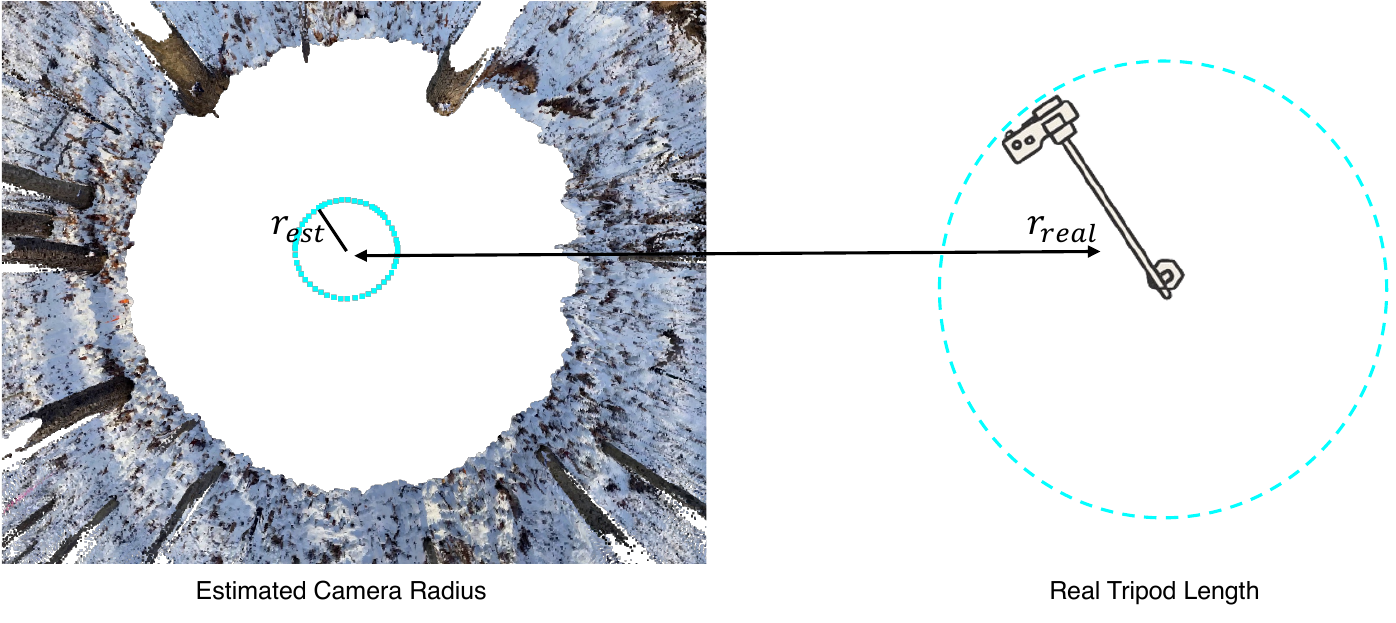}
\captionof{figure}{Metric scale recovery. \textit{Left}: top-down view of the reconstructed point cloud with the estimated camera trajectory radius $r_{\text{est}}$ (cyan). \textit{Right}: the known physical tripod arm length $r_{\text{real}}$ and the corresponding circular scanning path. The ratio $s = r_{\text{real}} / r_{\text{est}}$ converts all reconstructed quantities from the SLAM coordinate system to metric units.\label{fig:scale_recovery}}
\end{center}

\begin{figure*}[!h]
\centering
\includegraphics[width=.9\textwidth]{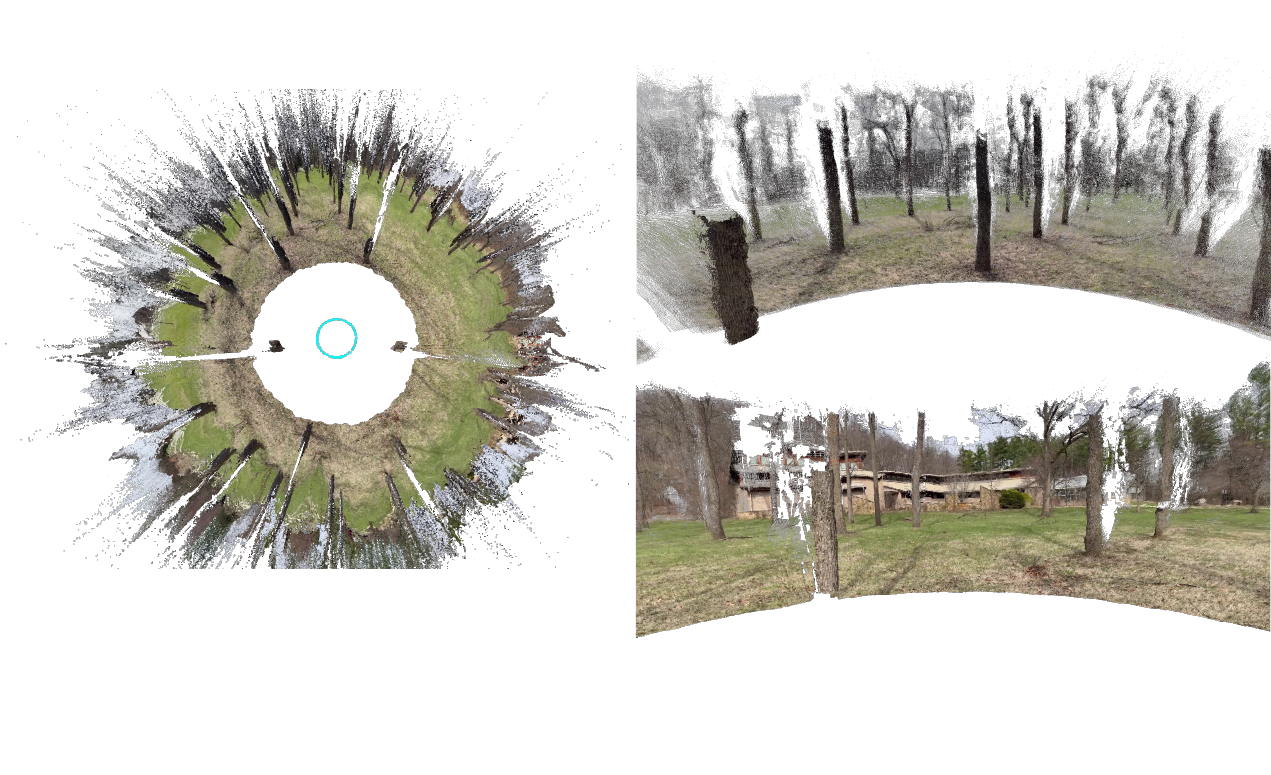}
\caption{Dense point cloud reconstruction from a single monocular video. \textit{Left}: top-down view showing the recovered camera trajectory (cyan ellipse) and surrounding tree stems. \textit{Right}: two front-view perspectives of the same reconstruction.\label{fig:slam_reconstruction}}
\end{figure*}

Figure~\ref{fig:slam_reconstruction} shows a representative reconstruction result for one of the evaluated plots. The top-view rendering (left) confirms that the circular camera trajectory (cyan circle) is accurately recovered, with reconstructed stems distributed around the plot center in a spatially consistent arrangement. The white central region corresponds to the unobserved interior of the scanning circle, which is expected given the outward-facing camera trajectory. The two front-view renderings (right) show the same reconstruction from different viewpoints, illustrating that individual tree stems are clearly resolved with coherent surface geometry recovered along the full visible trunk height, and that background structures including the ground plane and distant vegetation are faithfully captured.

\section{Results}\label{sec:results}
We evaluate our system in both plantation and natural forest environments at Purdue University’s Martell Forest. In the plantation, trees are arranged in a regular grid and are well maintained, with clear ground conditions. The plantation also provides accurate ground-truth measurements of tree DBHs and locations collected by forestry specialists. In the natural forest, trees are distributed irregularly and accompanied by naturally grown underbrush. Ground-truth data in the natural forest are measured manually by type, and accurate tree locations are unavailable. Although our system should theoretically operate similarly in both forest types, we report the results separately because of differences in the available ground-truth data.


\subsection{Plantation Dataset }


\begin{figure*}[t]
\centering

\begin{minipage}{0.62\linewidth}
\centering
\includegraphics[width=\linewidth]{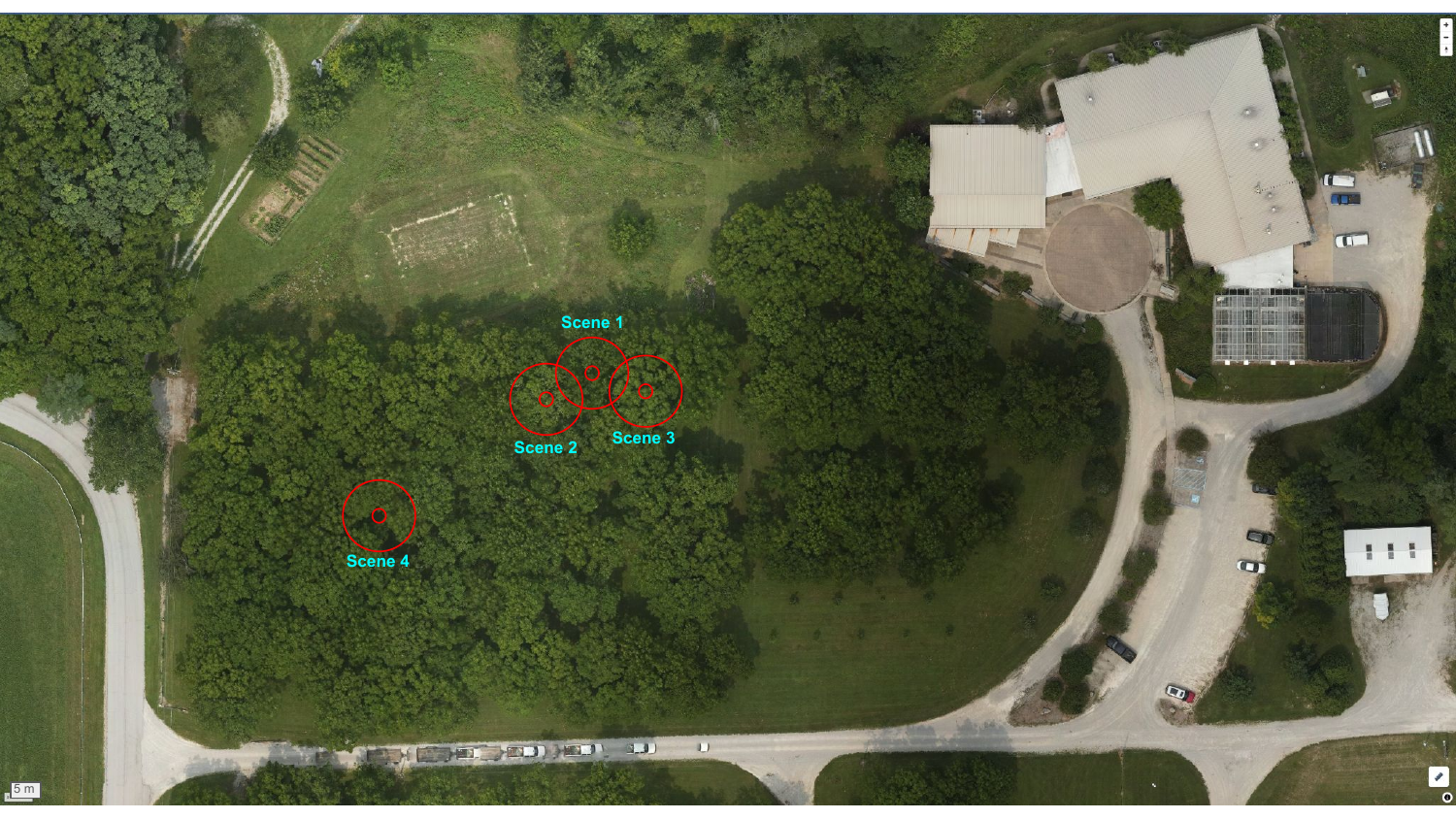}
\end{minipage}
\hfill
\begin{minipage}{0.36\linewidth}
\centering
\includegraphics[width=\linewidth]{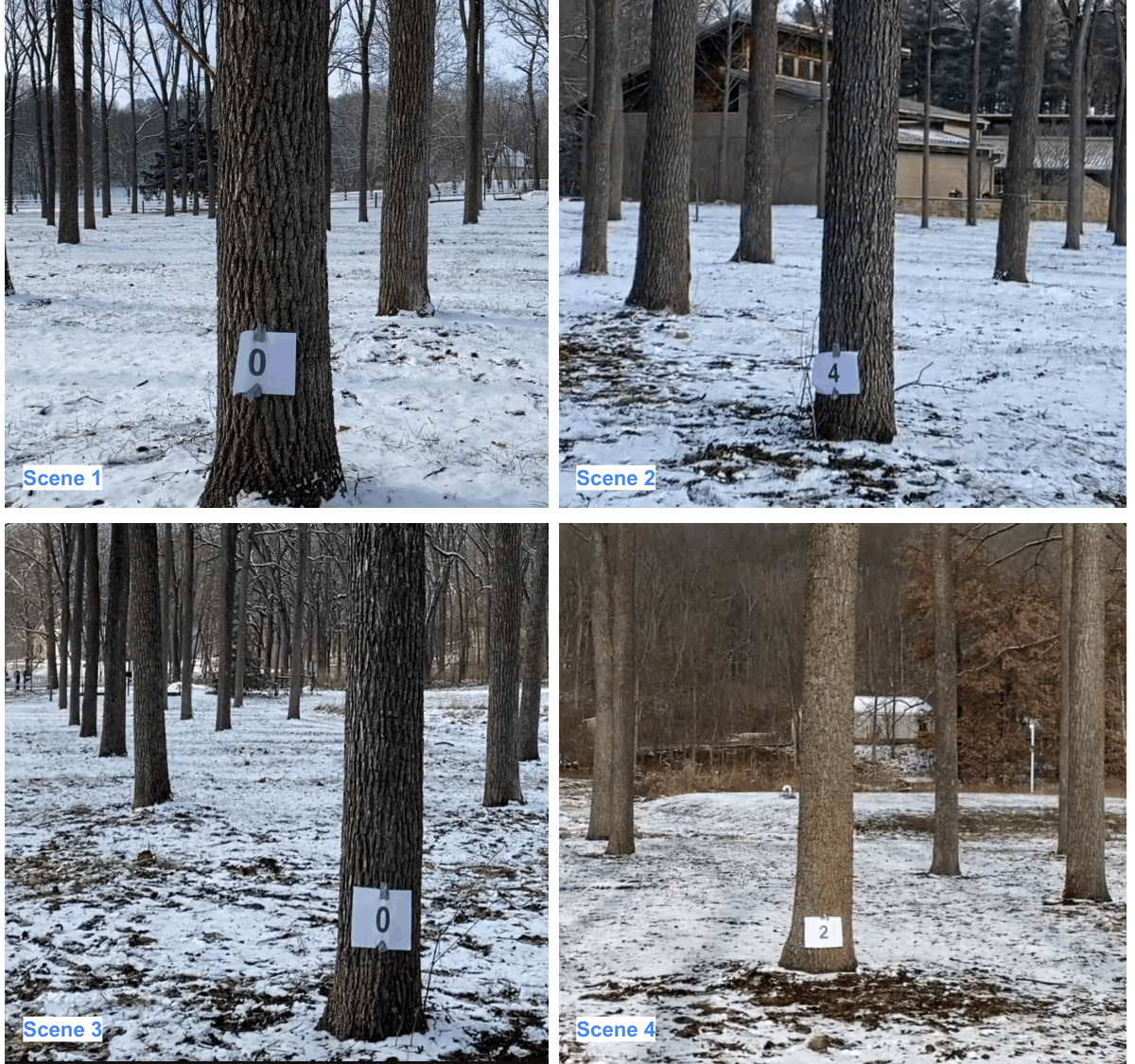}
\end{minipage}

\caption{
\textbf{Left:} Spatial distribution of the four circular sample plots used for system evaluation within Martell Forest, Purdue University. Each red circle indicates a circular inventory plot where smartphone-based circular video acquisition was performed (Scene~1--Scene~4, radius $\sim$10\,m). 
\textbf{Right:} Examples of field data collection. Trees were labeled with numbered markers to ensure consistent identity across videos and ground-truth DBH measurements, which were manually measured at breast height using a forestry diameter caliper.
}
\label{fig:circular_dataset_combined}
\end{figure*}


To evaluate the proposed system, data were collected from four circular sample plots located in Martell Forest at Purdue University. The study site represents a typical mixed forest environment with moderate tree density and understory vegetation. 

For ground truth construction, trees within each plot were manually identified and labeled using numbered paper markers attached to the tree trunks. 
Ground-truth DBH values were measured manually in the field using a forestry diameter caliper (tree diameter gauge) at breast height. The measured DBH values were recorded together with the corresponding tree ID numbers, establishing a direct correspondence between the physical trees observed in the videos and the ground-truth DBH measurements used for quantitative evaluation. Figure~\ref{fig:circular_dataset_combined} left part illustrates the field data collection process. Numbered markers are attached to tree trunks to visually identify each tree, allowing the recorded videos to be aligned with the corresponding DBH measurements obtained during field inventory.


At each plot, five video sequences were recorded using the circular acquisition setup described in the \textit{Methods} section. Three videos were captured from the same position but initiated from different starting directions in order to evaluate measurement repeatability under identical acquisition geometry. In addition, two videos were recorded from positions approximately 1~m offset from the plot center. These offset recordings simulate small positioning deviations that may occur during practical field deployment and allow us to assess the robustness of the system under changes in acquisition location.

All videos were recorded using a consumer smartphone camera at a resolution of 1080p and 30~FPS. In total, the dataset consists of 20 video sequences across the four plots. The dataset therefore provides multiple observations of the same trees under slightly different acquisition conditions, enabling evaluation of both DBH measurement accuracy and cross-video repeatability.

\begin{figure*}[!t]
\centering
\includegraphics[width=0.7\linewidth]{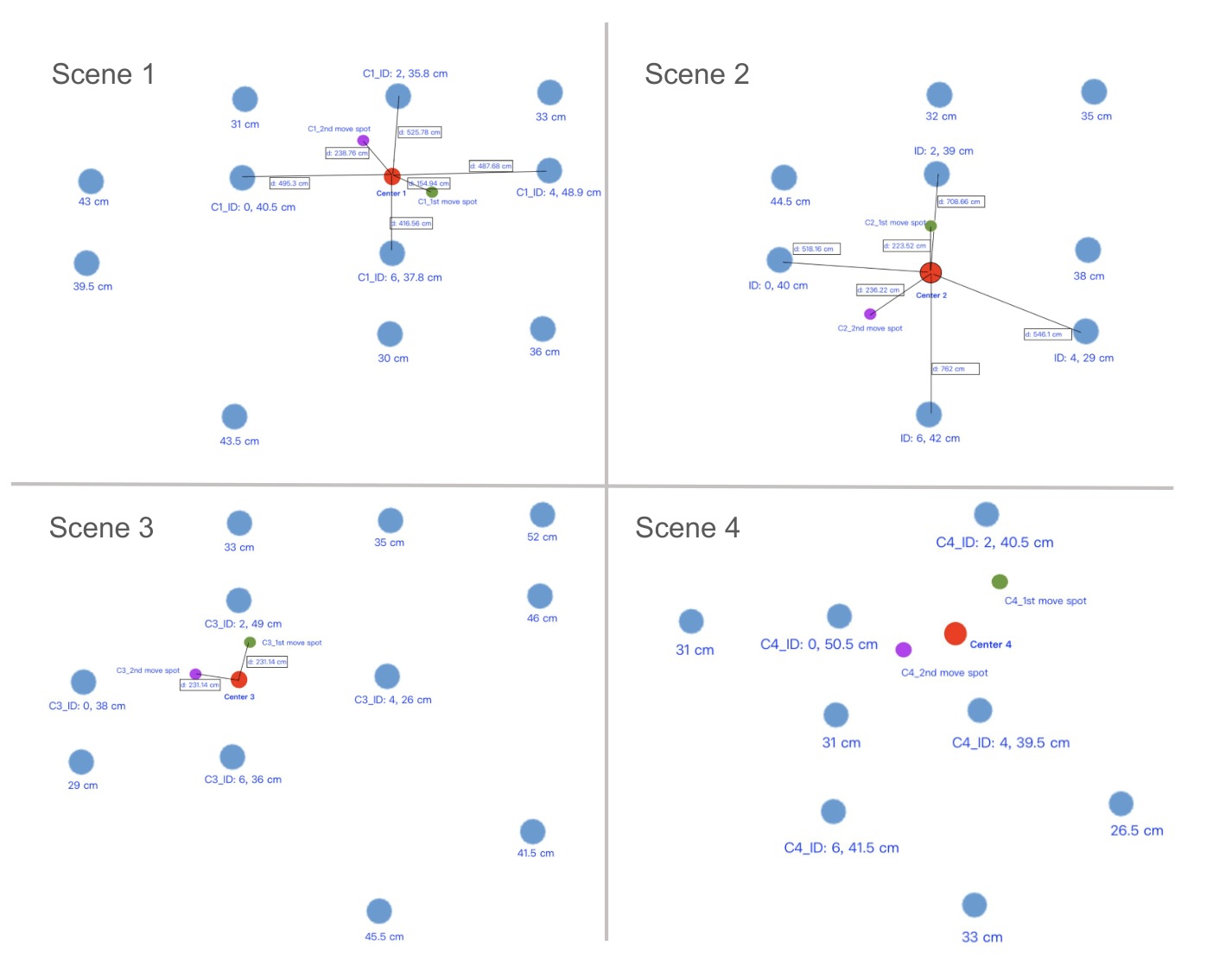}
\caption{Spatial layouts of the four evaluation plots. Blue points represent trees with ground-truth DBH measurements. The red marker indicates the primary position used for circular smartphone video acquisition. Green and purple markers denote additional offset acquisition positions used to evaluate repeatability under small variations in camera placement.}
\label{fig:plot_layout}
\end{figure*}

Figure~\ref{fig:plot_layout} shows the spatial layouts of the four evaluation plots in a plantation forest. In these diagrams, blue points represent tree locations with ground-truth DBH measurements. The positions of these trees were obtained through field measurements using GNSS combined with LiDAR-based tree mapping to provide accurate spatial references of the stems within each plot. The red marker indicates the primary position used for circular smartphone video acquisition. Moreover, the green and purple markers denote additional offset acquisition positions used to evaluate repeatability under small variations in camera placement. These offset positions are located approximately 1~m from the primary position and were introduced to simulate realistic deviations that may occur when operators place the acquisition stand in the field. By comparing measurements obtained from these slightly shifted positions, we can assess the robustness and repeatability of the proposed system under small changes in acquisition geometry.




In addition, the acquisition regions of Scene~1 to Scene~3 partially overlap, allowing several trees to be observed from different viewing angles across multiple video sequences. This overlap enables cross-scene consistency analysis, where the same trees are reconstructed and measured from different viewpoints. By comparing the DBH estimates obtained from these overlapping observations, we can further evaluate the stability of the proposed method under varying viewing geometries.

Overall, this dataset design provides multiple independent observations of the same trees under slightly different acquisition positions and viewing angles, enabling a comprehensive evaluation of both measurement accuracy and repeatability.

Table~\ref{tab:dataset} summarizes the main characteristics of each plot, including the number of trees with ground-truth DBH measurements and the statistical distribution of DBH values within the plot boundary.

\begin{table*}[!h]
\centering
\caption{Summary of the four evaluation plots. Tree count refers to the number of stems with ground-truth DBH measurements within the plot boundary.}
\label{tab:dataset}
\begin{tabular}{lcccc}
\hline
 & Scene 1 & Scene 2 & Scene 3 & Scene 4 \\
\hline
Number of trees     & 11 & 12 & 12 & 12 \\
DBH range (cm)      & 30.0-48.90 & 27.0-44.5 & 29.0-52.0 &26.5-50.5 \\
Mean DBH (cm)       & 38.09 & 36.04 & 38.79 & 36.70 \\
Videos per plot     & 5 & 5 & 5 & 5 \\
\hline
\end{tabular}
\end{table*}

\subsection{Implementation Details}
For monocular depth estimation, we use the Video-Depth-Anything-Base model \cite{zhao2024_video_depth_anything}, which produces temporally consistent dense depth maps from video input. For semantic tree segmentation, we adopt the deep learning-based tree detection and diameter estimation model of Grondin et al.\ \cite{10.1093/forestry/cpac043}, pretrained on the UTree photorealistic virtual forest benchmark \cite{10536353}. Both models are kept frozen throughout the pipeline without any further fine-tuning. The SLAM module and measurement pipeline run on a single NVIDIA RTX 4090 GPU. All experiments use the same hyperparameters across all four plots without scene-specific tuning.

\subsection{DBH Estimation Accuracy}

Table~\ref{tab:dbh_overall} summarizes the overall DBH estimation accuracy across all four plots. Our method achieves a mean absolute error (MAE) of 1.51\,cm, a root mean square error (RMSE) of 1.92\,cm, and a mean absolute relative error (MARE) of 3.98\%. The close agreement between MAE and RMSE further indicates that large outlier errors are uncommon and that the error distribution is well-behaved.

\begin{table}[!h]
\centering
\caption{Overall DBH estimation accuracy across all plots.\label{tab:dbh_overall}}
\begin{tabular*}{\columnwidth}{@{\extracolsep{\fill}}cccc}
\hline
MAE (cm) & RMSE (cm) & SD (cm) & MARE (\%) \\
\hline
1.51 & 1.92 & 1.86 & 3.98 \\
\hline
\end{tabular*}
\end{table}

\begin{figure*}[!h]
\centering
\includegraphics[width=\linewidth]{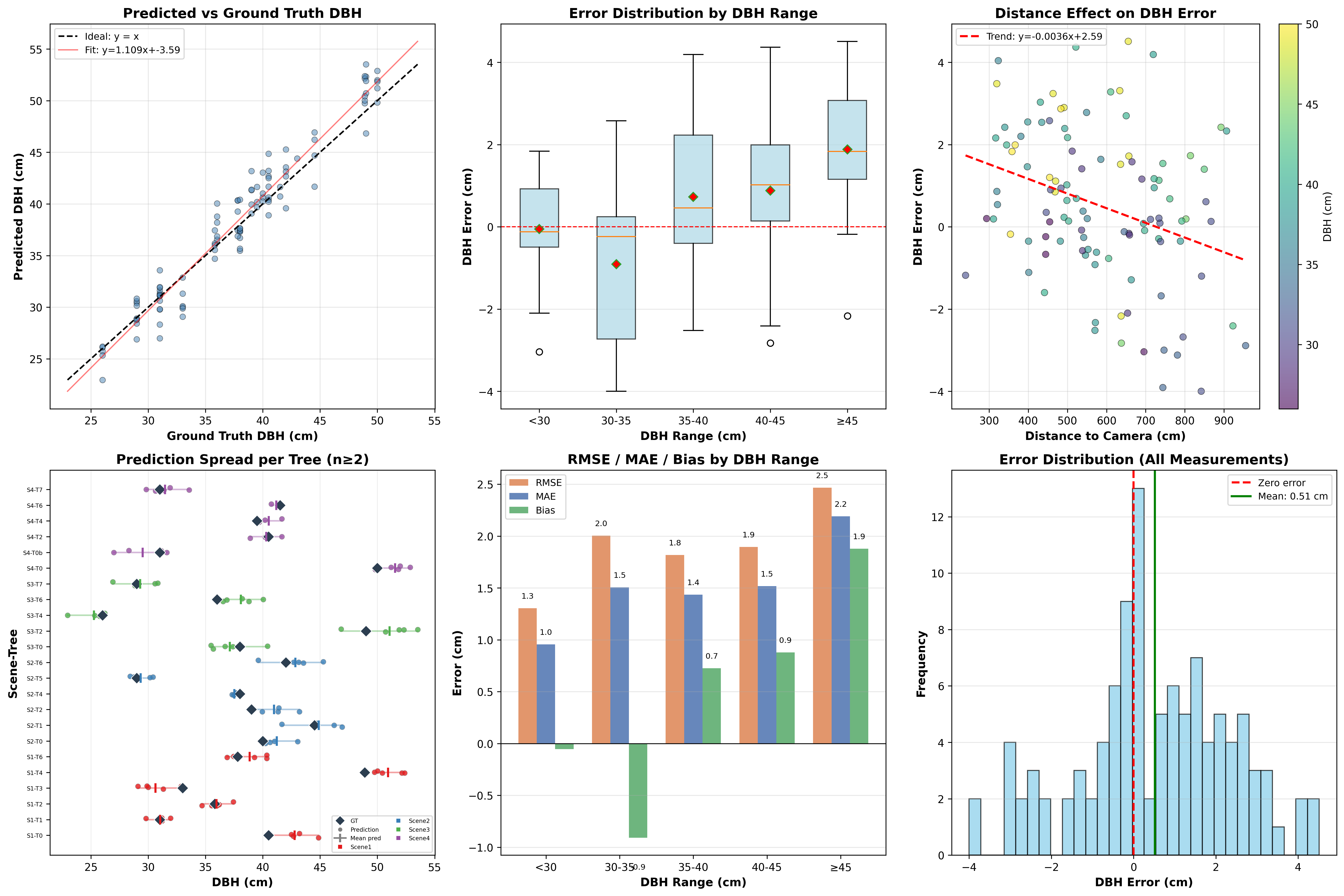}

\caption{Comprehensive DBH error analysis. \textit{Top row, left to right}: predicted vs.\ ground truth scatter plot with linear fit; error distribution stratified by DBH range; effect of camera-to-stem distance on DBH error, colored by stem size. \textit{Bottom row, left to right}: per-tree prediction spread across videos for each scene; RMSE, MAE, and bias stratified by DBH range; overall error histogram with mean indicated.\label{fig:error_analysis}}
\end{figure*}

The error distribution confirms the reliability of individual estimates: 42.45\% of measurements fall within 1\,cm of ground truth, 66.03\% within 2\,cm, and only 13.21\% exceed 3\,cm (Table~\ref{tab:error_dist}). The overall error histogram in Figure~\ref{fig:error_analysis} (bottom right) shows a distribution centered near zero and no heavy tails, indicating the absence of systematic large-scale failures.

\begin{table}[!h]
\centering
\caption{DBH error distribution across all measurements.\label{tab:error_dist}}
\begin{tabular}{cc}
\hline
Error Range & Proportion \\
\hline
$< 1$\,cm    & 42.45\% \\
$1$--$2$\,cm & 23.58\% \\
$2$--$3$\,cm & 20.75\% \\
$\geq 3$\,cm & 13.21\% \\
\hline
\end{tabular}
\end{table}

The predicted vs.\ ground truth scatter plot (Figure~\ref{fig:error_analysis}, top left) shows strong linear agreement across the full range of stem sizes ($R^2 = 0.964$), with a fitted slope of 1.109 and intercept of $-3.59$\,cm. The error distribution by DBH range (Figure~\ref{fig:error_analysis}, top center) confirms that the median error remains close to zero across all size classes, indicating the absence of systematic bias. Estimation variability increases moderately for stems above 40\,cm, which is expected as larger trunks produce more reconstructed points near the foreground-background boundary where depth estimates are less reliable. Nevertheless, even for the largest stems the interquartile range remains within $\pm$2\,cm. The RMSE/MAE/bias breakdown (Figure~\ref{fig:error_analysis}, bottom center) further confirms that MAE stays below 2.2\,cm across all DBH ranges, demonstrating robust performance regardless of stem size. The per-tree prediction spread (Figure~\ref{fig:error_analysis}, bottom left) shows that estimates from all five videos remain tightly grouped around the ground truth for the large majority of trees across all four scenes, reflecting strong cross-video repeatability at the individual tree level.

The error distribution by DBH range (Figure~\ref{fig:error_analysis}, top center) shows that the median error remains close to zero for smaller stems ($<$30\,cm) but shifts upward for larger ones, consistent with the overestimation trend described above. The interquartile range also widens moderately with increasing stem size, reflecting greater depth uncertainty at the edges of larger trunks.

The distance effect panel (Figure~\ref{fig:error_analysis}, top right) shows a weak negative trend ($y = -0.0036x + 2.59$) between camera-to-stem distance and DBH error. However, the slope is near zero and the scatter is large, indicating that distance has negligible influence on estimation accuracy within the range of our plots.    The slight trend may partly reflect the fact that closer trees are observed across more frames during the circular scan, increasing the number of reconstructed points but also accumulating more noisy edge points that widen the projected cross-section. In contrast, more distant trees are captured in fewer frames with a narrower angular extent, producing a tighter point distribution that is less susceptible to edge noise. 

\subsection{Per-Scene Analysis}
Per-scene results are reported in Table~\ref{tab:dbh_perscene} and visualized in Figure~\ref{fig:error_analysis}. MARE ranges from 3.26\% (Scene 4) to 5.11\% (Scene 3), and the error boxplots by scene (Figure~\ref{fig:error_analysis}, bottom left) show that all four scenes have median errors near zero with comparable spread. Scene 3 exhibits a slightly wider interquartile range and a higher incidence of larger errors, which may reflect more complex understory conditions at that site. Importantly, no scene shows a systematic negative bias, and the inter-scene variation is modest, indicating that the method generalizes consistently across different plot configurations without scene-specific tuning.

\begin{table}[!h]
\centering
\caption{Per-scene DBH estimation accuracy.\label{tab:dbh_perscene}}
\begin{tabular}{lccc}
\hline
Scene & MAE (cm) & RMSE (cm) & MARE (\%) \\
\hline
Scene 1 & 1.58 & 2.00 & 4.10 \\
Scene 2 & 1.35 & 1.75 & 3.42 \\
Scene 3 & 1.88 & 2.24 & 5.11 \\
Scene 4 & 1.22 & 1.61 & 3.26 \\
\hline
\end{tabular}
\end{table}

The per-tree prediction spread (Figure~\ref{fig:error_analysis}, bottom left) illustrates DBH estimates from all five videos against ground truth for each scene, grouped by scene and tree index. Across all four scenes, the five video estimates remain tightly clustered around the ground truth for the large majority of trees, demonstrating strong cross-video repeatability at the individual tree level. The estimates from different videos are mutually consistent within each scene, confirming that the pipeline produces stable measurements regardless of starting direction.

\subsection{Tree Localization and Cross-Video Consistency}
\begin{figure*}[!h]
\centering
\includegraphics[width=\linewidth]{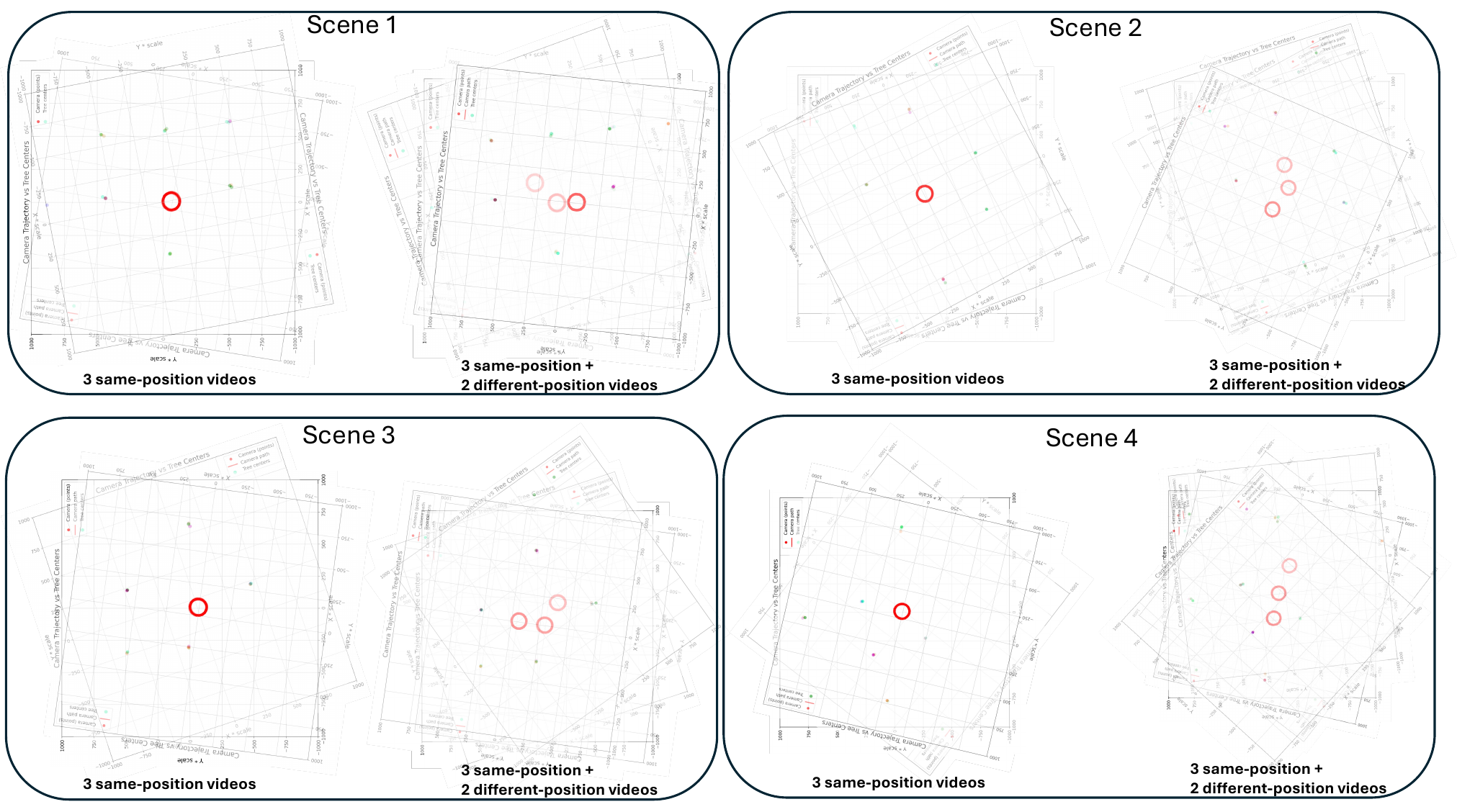}
\caption{Cross-video tree localization consistency across the four plots. For each scene, the left panel overlays predicted tree positions from the three same-center videos, and the right panel overlays all five videos including two captured from offset positions. Colored dots represent predicted tree centers from individual videos; overlapping clusters indicate consistent localization.} 
\label{fig:localization}
\end{figure*}
In forest inventory, repeatable tree localization is essential for constructing reliable plot maps and enabling consistent revisits over time. Since independent centimeter-level ground truth for tree positions was not available, we evaluate localization quality indirectly through cross-video consistency: stable and overlapping tree position maps across independently captured videos of the same plot provide strong evidence that the estimated positions faithfully reflect the true spatial arrangement of stems.

For each plot, we generated two types of overlay maps: one overlaying the predicted positions from the three videos captured at the same position (but starting from different directions), and one overlaying all five videos including the two captured from an offset position. These overlays are shown in Figure~\ref{fig:localization}.

Across all four scenes, the three same-center videos produce tree position maps that align closely, with predicted stem locations consistently clustering into tight, well-separated groups. This demonstrates that the pipeline recovers stable absolute positions regardless of which tree the operator began the rotation from, confirming robustness to variation in starting conditions. The all-five-video overlay shows similarly tight clustering for most stems. For a subset of trees near the plot boundary, slightly greater spread is visible when the offset videos are included, which is expected, given that the different positions alter the effective viewing geometry and scanning radius for peripheral stems. Nonetheless, the relative arrangement and approximate positions of all detected stems are preserved consistently across all five videos in every scene.

\subsection{Evaluation on Natural Forest}
To assess generalization beyond the managed plantation, we evaluate our system on a natural forest plot exhibiting irregular stem distribution, dense understory, and uneven ground. Seven videos were acquired under two protocols: V1--V3 share the same center but start from different rotational directions, isolating sensitivity to initialization; V4--V7 are captured from four distinct centers within the same plot, probing robustness to viewpoint variation. Each video covers an overlapping subset of stems, yielding multiple independent observations per tree.
 
\begin{figure*}[!h]
\centering
\includegraphics[width=\linewidth]{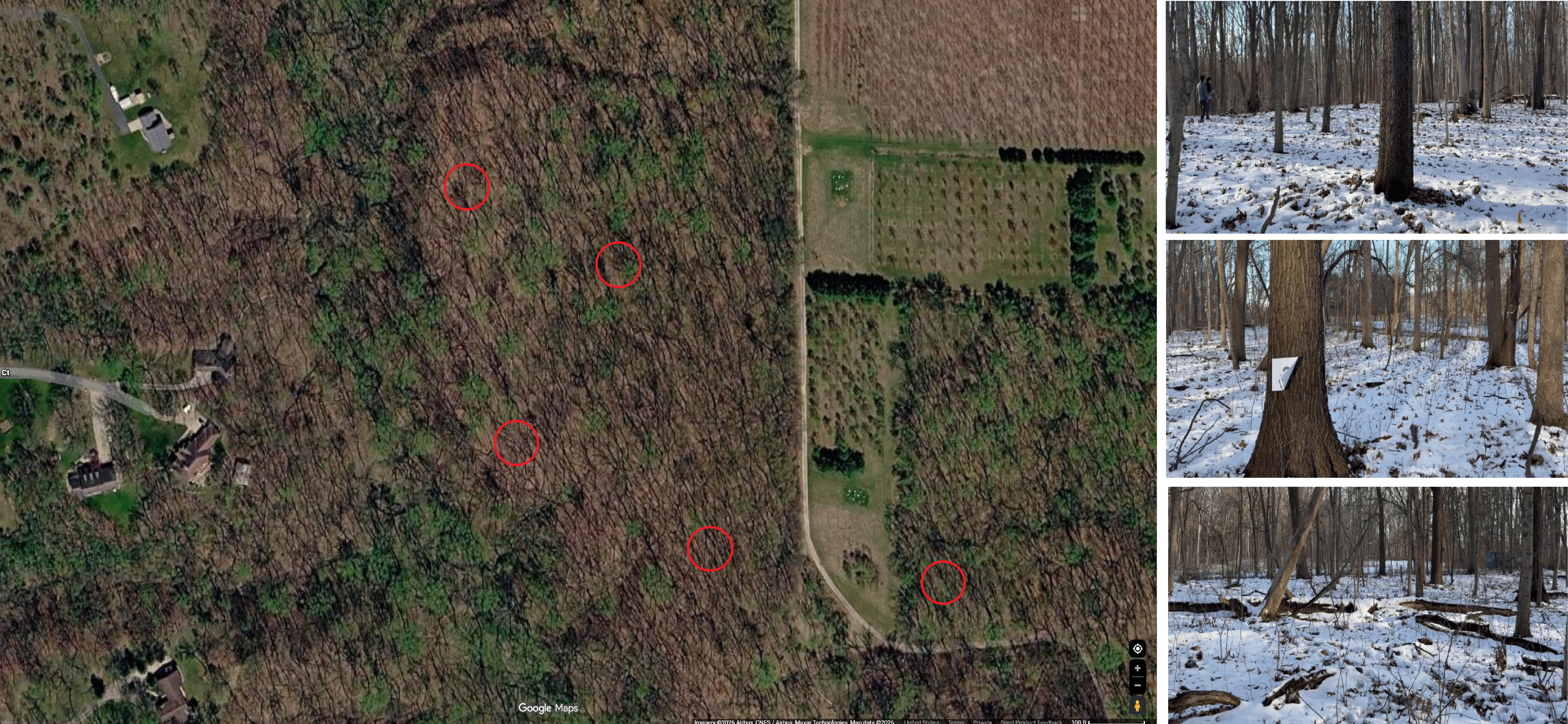}
\caption{Natural forest evaluation site at Martell Forest. \textbf{Left:} aerial overview with candidate sampling locations marked. \textbf{Right:} representative on-site views illustrating the stem distribution, understory density, and lighting conditions encountered during acquisition.\label{fig:natural_site}}
\end{figure*}
 
Aggregate accuracy is reported in Table~\ref{tab:natural_overall}. The system attains an MAE of 2.30\,cm and RMSE of 2.55\,cm, indicating that DBH estimates are essentially unbiased on this plot.
 
\begin{table}[!h]
\centering
\caption{Overall DBH estimation accuracy on the natural forest plot \label{tab:natural_overall}}
\begin{tabular*}{\columnwidth}{@{\extracolsep{\fill}}cccc}
\hline
MAE (cm) & RMSE (cm) & SD (cm) & MARE (\%) \\
\hline
2.30 & 2.55 & 2.58 & 5.69 \\
\hline
\end{tabular*}
\end{table}
 
Per-video results are summarized in Table~\ref{tab:natural_pervideo}. MAE varies from 1.41\,cm (V4) to 3.24\,cm (V6), and MARE from 3.66\% to 7.84\%. The same-center videos (V1--V3) show comparable error magnitudes despite different initialization, while the cross-center videos (V4--V7) span a similar range, indicating that the reconstruction is not strongly biased by the choice of acquisition center.
 
\begin{table}[!h]
\centering
\caption{Per-video DBH estimation accuracy on the natural forest plot. V1--V3 are recorded from a shared center; V4--V7 are recorded from four distinct centers within the same plot.\label{tab:natural_pervideo}}
\begin{tabular}{lcccc}
\hline
Video & MAE (cm) & RMSE (cm) & MARE (\%) \\
\hline
V1 & 2.20 & 2.41 & 5.33 \\
V2 & 1.64 & 2.00 & 3.80 \\
V3 & 3.10 & 3.15 & 7.81 \\
V4 & 1.41 & 1.71 & 3.66 \\
V5 & 2.36 & 2.57 & 6.35 \\
V6 & 3.24 & 3.33 & 7.84 \\
V7 & 1.84 & 2.08 & 4.27 \\
\hline
\end{tabular}
\end{table}
 
\begin{figure*}[!h]
\centering
\includegraphics[width=\linewidth]{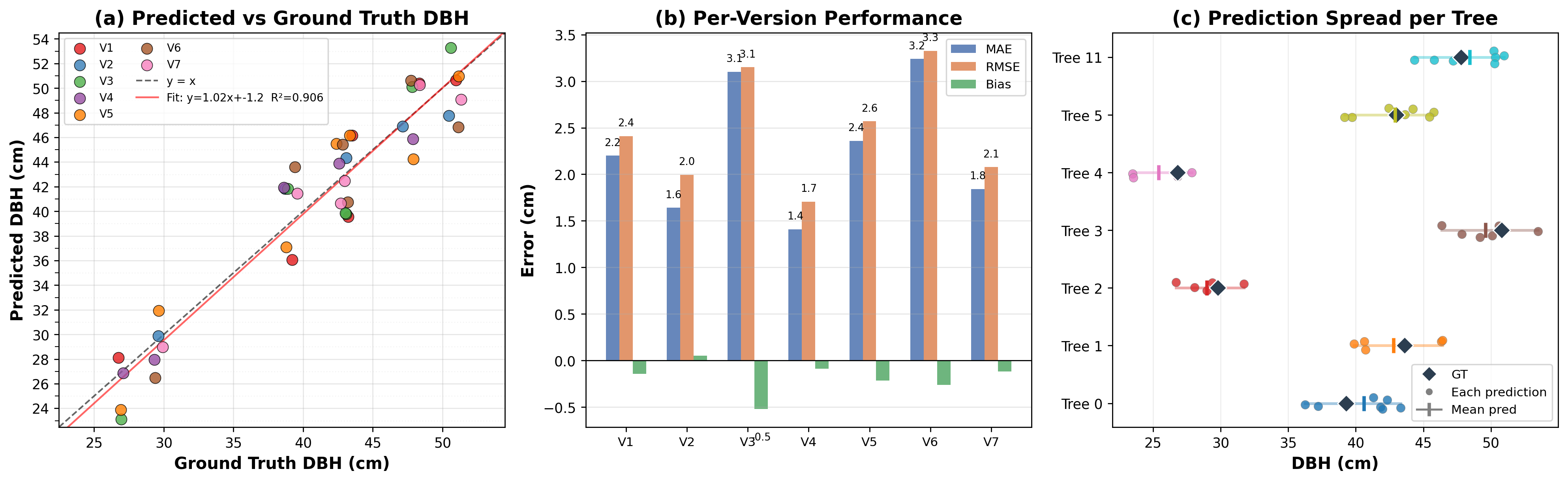}
\caption{DBH estimation analysis on the natural forest plot. (a) Predicted vs.\ ground truth DBH across seven videos, with linear fit ($y=1.02x-1.2$, $R^2=0.906$). (b) Per-video MAE, RMSE, and bias. (c) Per-tree prediction spread: individual estimates (circles), mean prediction (cross), and ground truth (diamond).\label{fig:natural_results}}
\end{figure*}

Figure~\ref{fig:natural_results} summarizes the dataset-level behavior. The predicted--ground-truth scatter (a) shows strong linear agreement, with a least-squares fit of $y = 1.02x - 1.2$ and $R^2 = 0.906$, indicating near-unit slope and negligible global offset. The per-video breakdown (b) confirms that bias magnitudes remain below 0.6\,cm for all videos and that MAE/RMSE values are concentrated in the 1.4--3.3\,cm range. The per-tree spread (c) shows that, for most stems, predictions from different videos cluster tightly around the ground truth; the residual spread is largest for a few stems where partial occlusion from understory limits the visible bark surface.
 
\begin{figure*}[!h]
\centering
\includegraphics[width=\linewidth]{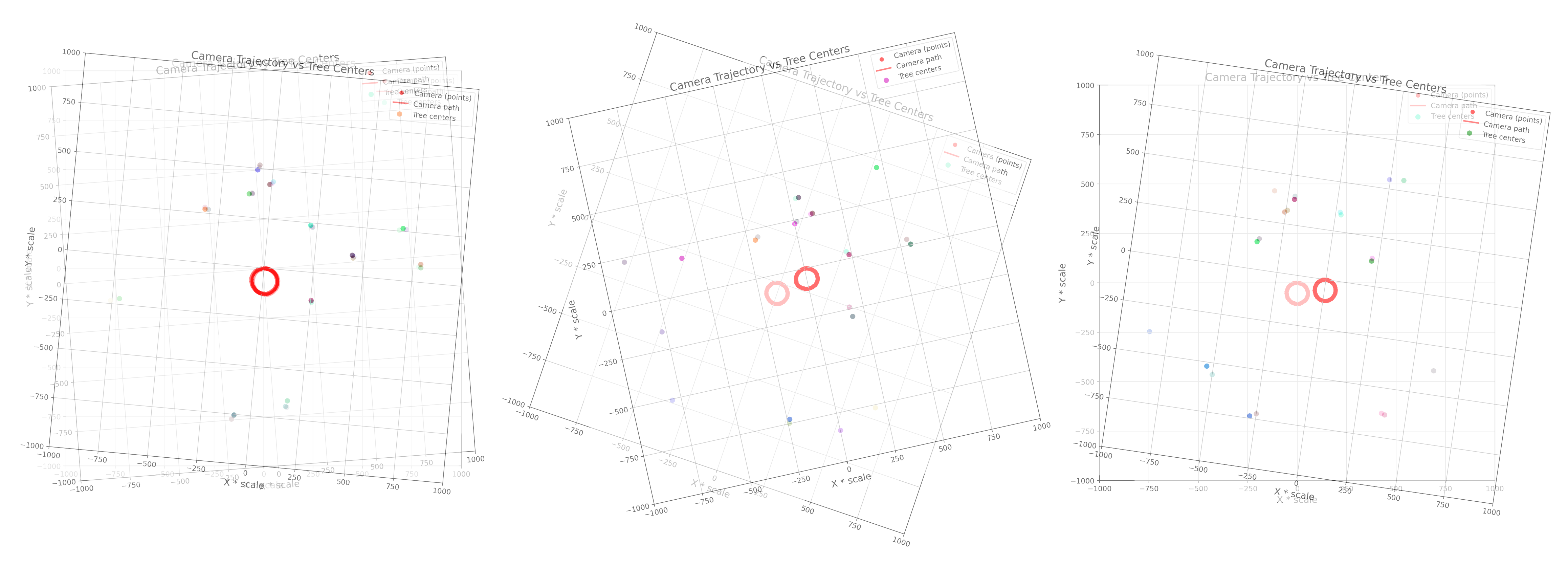}
\caption{Tree position maps recovered on the natural forest plot. \textbf{Left:} overlay of the three same-center videos initiated from different directions. \textbf{Middle, Right:} overlays of pairs of videos captured from different centers within the same plot. Red circles denote estimated camera trajectories and colored dots denote detected tree positions. Stem arrangements are consistent across panels under both initialization and viewpoint variation.\label{fig:natural_localization}}
\end{figure*}

Figure~\ref{fig:natural_localization} overlays the recovered tree position maps under both acquisition protocols. The left panel superimposes the three same-center videos starting from different directions, in which a single trajectory is observed and stem positions cluster tightly across runs, indicating that localization is insensitive to starting direction. The middle and right panels each overlay two videos captured from different centers within the same plot; the trajectories appear as separate circles offset by the center displacement, while the recovered stem positions remain mutually consistent up to that offset. Across all three panels the relative arrangement of stems is preserved, confirming robust SLAM-based localization under both initialization and viewpoint variation.
 
\subsection{Runtime}
 
We report the end-to-end processing time of our pipeline on a single NVIDIA RTX 4090 Ti GPU. For a typical one-minute input video (1080p at 30\,FPS, approximately 1800 frames), the full pipeline including monocular depth estimation, SLAM reconstruction, segmentation, and DBH measurement completes in approximately 5 minutes. Table~\ref{tab:runtime} compares the total field-to-result time of our method against manual measurement and traditional SfM-based photogrammetry.
 
\begin{table*}[!h]
\centering
\caption{Comparison of field acquisition and processing time for plot-level DBH estimation.\label{tab:runtime}}
\begin{tabular}{lccc}
\hline
Method & Field time & Processing time & Total \\
\hline
Manual tape measurement     & 15--30 min & --             & 15--30 min \\
SfM-MVS          & 10--20 min & 30--60 min     & 40--80 min \\
\textbf{Ours}               & \textbf{$<$2 min} & \textbf{$\sim$5 min} & \textbf{$\sim$7 min} \\
\hline
\end{tabular}
\end{table*}

Manual tape measurement requires no computation but scales poorly with plot size due to per-tree handling time. SfM-based photogrammetry requires the operator to capture many overlapping images from multiple viewpoints around the plot, followed by computationally expensive feature matching and dense reconstruction that can take 30--60 minutes depending on the number of input images \cite{8588151}. Our system reduces both field and processing time substantially: a single one-minute video replaces multi-path image acquisition, and the learning-based SLAM pipeline avoids the costly exhaustive feature matching required by conventional SfM.

\section{Discussion}\label{sec:discussion}
Our experiments show that a consumer smartphone, combined with a circular video acquisition protocol and a pipeline that couples video depth priors, monocular SLAM, and segmentation-guided measurement, achieves an MAE of 1.51\,cm (3.98\% MARE) in DBH measruement across four managed plots and 2.30\,cm MAE (5.69\% MARE) on a structurally more complex natural forest plot. The aggregate fit on the natural plot ($y = 1.02x - 1.2$, $R^2 = 0.906$) and the near-zero mean error ($-0.19$\,cm) indicate that estimates are essentially unbiased at the dataset level.Smartphone photogrammetry pipelines report RMSE in the 1--3\,cm range depending on stem size and occlusion~\cite{ahamed2023_phone_photogrammetry}, and LiDAR-equipped tablet applications can reach sub-centimeter agreement on individual stems~\cite{howie2024_ipad_lidar_dbh} at the cost of specialized hardware. Our system operates within this accuracy envelope while relying only on a standard smartphone and a single short capture.

The magnitude of per-video error varies across the natural plot (MAE 1.4--3.2\,cm) without a corresponding shift in mean error, suggesting that residuals are dominated by stem-level reconstruction noise rather than a systematic scale or offset bias. The largest per-tree spreads occur on stems where understory vegetation partially occludes the trunk surface, reducing the number of valid points contributing to the cross-section fit. Filtering the projected slice to suppress mixed-depth points near the silhouette boundary is a natural direction for tightening the per-video distribution.

Cross-video analysis provides evidence of reliability that DBH error metrics alone do not capture. On the natural plot, V1--V3 share a center and differ only in starting direction; their error distributions are comparable, indicating that the reconstruction is largely insensitive to initialization. V4--V7 are recorded from four distinct centers within the same plot and span a similar error range, showing that no single viewpoint dominates the result and that the pipeline is robust to the choice of acquisition center. On the managed plots, the slightly larger positional spread observed for boundary stems under offset acquisition is geometrically expected: peripheral stems are viewed from more oblique angles and at greater range, reducing point density on the visible bark. This boundary effect reflects a property of the circular acquisition geometry rather than a failure of the reconstruction.

Treating Video Depth Anything as a soft prior with learnable per-frame scale and offset, rather than as a hard supervisory signal, is motivated by the low-texture nature of forest scenes, in which purely geometric SLAM produces noisier reconstructions, and by the scale and offset inconsistencies inherent in general-purpose monocular depth models. The soft formulation lets the optimization exploit the structural cues in the depth predictions without propagating their absolute calibration errors into geometry or trajectory. This design may transfer to other outdoor reconstruction settings where domain-specific depth models are unavailable. The circular trajectory further provides natural loop-closure opportunities that suppress drift over the full 360\textdegree{} rotation.

Several extensions remain. Estimating tree height from the reconstructed point cloud would enable derived attributes such as stem volume and above-ground biomass without additional sensing. Joint species classification from the captured frames would increase the information yield of a single pass. Evaluation on denser stands, tropical environments, and leaf-on conditions will be necessary to characterize operational limits. Finally, lightweight domain adaptation of the depth prior on forest-specific data may further reduce per-stem variance and tighten the residual distribution observed on natural plots.

\section{Conclusions}\label{sec:conclusions}
We presented a smartphone-based pipeline for circular plot forest inventory that estimates tree DBH and position from a single handheld monocular video. By combining a video depth prior, dense monocular SLAM with loop closure, and a segmentation-guided measurement module, the system achieves 1.51\,cm MAE and 3.98\% MARE for DBH estimation across four managed forest plots, and 2.30\,cm MAE and 5.69\% MARE on an additional natural forest plot with greater structural complexity. Across both datasets, the system demonstrates stable and repeatable tree localization across independently captured videos from different starting directions and positions. The approach requires no specialized hardware beyond a consumer smartphone and a simple stand, and completes data collection in under two minutes per plot with approximately five minutes of processing time. These results suggest that modern computer vision techniques can bridge the gap between the accuracy of established inventory methods and the accessibility demands of large-scale, low-cost forest monitoring.

\section*{Data Availability}
The codes and datasets of the paper are available from the corresponding author on reasonable request.

\section*{Declaration of Competing Interest}
The authors declare no conflicts of interest.

\section*{Funding}
This work is based upon efforts supported by the PERSEUS grant, \#2023-68012-38992 under USDA NIFA. The views and conclusions contained herein are those of the authors and should not be interpreted as representing the official policies, either expressed or implied, of NIFA or the U.S. Government. The U.S. Government is authorized to reproduce and distribute reprints for governmental purposes notwithstanding any copyright annotation therein.

\printcredits

\bibliographystyle{cas-model2-names}

\bibliography{references}

\end{document}